# Text Classification using the Concept of Association Rule of Data Mining

**Chowdhury Mofizur Rahman**
Department of Computer Science
Stamford University Bangladesh
cmrbuet@yahoo.com

**Ferdous Ahmed Sohel**
Department of Computer Science & Engineering
International Islamic University Chittagong, Dhaka Campus
ferasohel27@yahoo.com

**Parvez Naushad**
Department of Computer Science & Engineering
Bangladesh University of Engineering & Technology

**S M Kamruzzaman**
Department of CSE
International Islamic University Chittagong

## Abstract

As the amount of online text increases, the demand for text classification to aid the analysis and management of text is increasing. Text is cheap, but information, in the form of knowing what classes a text belongs to, is expensive. Automatic classification of text can provide this information at low cost, but the classifiers themselves must be built with expensive human effort, or trained from texts which have themselves been manually classified. In this paper we will discuss a procedure of classifying text using the concept of association rule of data mining. Association rule mining technique has been used to derive feature set from pre-classified text documents. Naïve Bayes classifier is then used on derived features for final classification.

## 1. Introduction

Text classification is the automated assigning of natural language texts to predefined categories based on their content. Text Classification is the primary requirement of Text Retrieval systems, which retrieve texts in response to a user query, and Text Understanding systems, which transform text in some way such as producing summaries, answering questions or extracting data.

There exist some algorithms for learning to classify text based on the Naïve Bayes classifier. The probabilistic approaches for learning to classify text are described by Lewis (Lewis et al. 1992). In applying Naïve Bayes classifier, each word position in a document is defined as an attribute and the value of that attribute to be the English word found in that position. Naïve Bayes classification is given by:

$$V_{NB} = \mathrm{argmax}\ P(V_j) \prod P(a_i | V_j)$$

To summarize, the Naïve Bayes classification $V_{NB}$ is the classification that maximizes the probability of observing the words that were actually found in the example documents, subject to the usual Naïve Bayes independence assumption. The first term can be estimated based on the fraction of each class in the training data. The following equation is used for estimating the second term:

$$\frac{n_k + 1}{n + |\text{vocabulary}|}$$

where $n$ is the total number of word positions in all training examples whose target value is $V_j$, $n_k$ is the number of times that word is found among these $n$ words positions, and $|\text{vocabulary}|$ is the total number of distinct words found within the training data.

The proposed system is given a set of example documents. We first preprocess the text documents by parsing and removing stop words (Frank). We then collect frequently occurring words from each document. Each document is treated as a transaction and the set of frequently occurring words are viewed as a set of items in the transaction. We then apply association mining method (Frank, 2000) to discover sets of associated words in the documents. These set of associated words act as features. We them classify new documents using Naïve Bayes approach but using derived feature sets.

### 2.1 Data Mining

Popularly referred to as Knowledge Discovery [4] in Databases (KDD), Data Mining is the automated

extraction of patterns representing knowledge implicitly stored in large databases, data warehouses, and other massive information repositories. Standard data mining methods may be integrated with information retrieval techniques and the construction or use of hierarchies specifically for text data as well as discipline-oriented term classification systems (such as in chemistry, medicine, law, or economics).

Text databases are databases that contain word descriptions for objects. These word descriptions are usually not simple keywords but rather long sentences or paragraphs, such as product specifications, error or bug reports, warning messages, summary reports, notes, or other documents. The widely used and well-known data mining functionalities are Characterization and Discrimination, content based analysis (Hayes, 1990), Association Analysis, Classification and Prediction (Han, 2001), Cluster Analysis (Lewis, 1990), Outlier Analysis, Evolution Analysis. For our text classification purpose we have used Association Analysis for generating associative word sets.

## 2.2 Association rule

Association rule mining finds interesting association or correlation relationships among a large set of data items. The discovery of interesting association relationships among huge amounts of transaction records can help in many decision making processes. Let us consider the following assumptions for representing the Association rule in terms of mathematical representation, **J** = {$i_1$, $i_2$, … , $i_m$} be a set of items. **D** = Set of database transactions where each transaction T is a set of items such that T $\subseteq$ J. Each transaction is associated with an identifier, called TID. **A, B** = Set of items. A transaction T is said to contain A if and only if A $\subseteq$ T. An association rule is an implication of the form A $\Rightarrow$ B, where A $\subset$ J, B $\subset$ J, and A $\cap$ B = $\varnothing$ The rule A $\Rightarrow$ B holds in the transaction set D with support *S*, where *S* is the percentage of transaction in D that contain A $\cup$ B, i.e., Support (A $\Rightarrow$ B) = P (A $\cup$ B). The rule A $\Rightarrow$ B has confidence *C* in the transaction set D if *C* is the percentage of transaction in D containing A that also contain B, i.e., confidence (A $\Rightarrow$ B) = P (B│A) = [support count(A $\cup$ B) / support count(A)].

We now define some of the terminologies. Rules that satisfy both a minimum support threshold (*min_sup*) and a minimum confidence threshold (*min_conf*) are called **strong**. A set of items is referred to as an **itemset**. An itemset that contains *k* items is a *k*-**itemset**. The **occurrence frequency of an itemset** is the number of transactions that contain the itemset. This is also known, simply, as the **frequency**, **support count**, or **count** of the itemset. An itemset satisfies minimum support if the occurrence frequency of the itemset is greater than or equal to the product of *min_sup* and the total number of transactions in *D*. The number of transactions required for the itemset to satisfy minimum support is therefore referred to as the **minimum support count**. If an itemset satisfies minimum support, then it is a **frequent** itemset.

## 2.3 The Apriori Algorithm

**Apriori** is an influential algorithm for mining frequent itemsets for Boolean association rules. The name of the algorithm is based on the fact that the algorithm uses prior knowledge of frequent itemset properties. Association rule mining is a two steps process.

1. **Find all frequent itemsets**: By definition, each of these itemsets will occur at least as frequently as a pre-defined minimum support count.
2. **Generate strong Association rules from the frequent itemsets**: By definition, these rules must satisfy minimum confidence.

**Apriori** employs an iterative approach known as a **level-wise** search, where *k*-itemsets are used to explore *(k+1)*-itemsets. First, the set of frequent 1-itemsets is found. This set is denoted $L_1$. $L_1$ is used to find $L_2$, the set of frequent 2-itemsets, which is used to find $L_3$, and so on, until no more frequent *k*-itemsets can be found. The finding of each $L_k$ requires one full scan of the database. An important property called **Apriori property**, based on the observation is that, if an itemset *I* is not frequent, that is, P(*I*) < min_sup then if an item *A* is added to the itemset *I*, the resulting itemset (i.e., *I* $\cup$ *A*) cannot occur more frequently than *I*. Therefore, *I* $\cup$ *A* is not frequent either, that is, P(*I* $\cup$ *A*) < *min_sup*. To understand how Apriori property is used in the algorithm, let us look at how $L_{k-1}$ is used to find $L_k$. A two-step process is followed, consisting of **join** and **prune** actions.

**The join step:** To find $L_k$, a set of candidate $k$-itemsets is generated by joining $L_{k-1}$ with itself. This set of candidates is denoted by $C_k$. Let $l_1$ and $l_2$ be itemsets in $L_{k-1}$ then $l_1$ and $l_2$ are joinable if their first $(k-2)$ items are in common, i.e., $(l_1[1]=l_2[1])$ . $(l_1[2]=l_2[2])$ …… $(l_1[k-2]=l_2[k-2])$ . $(l_1[k-1]<l_2[k-1])$.

**The prune step:** $C_k$ is the superset of $L_k$. A scan of the database to determine the count of each candidate in $C_k$ would result in the determination of $L_k$ (itemsets having a count no less than minimum support in $C_k$). But this scan and computation can be reduced by applying the Apriori property. Any $(k-1)$-itemset that is not frequent cannot be a subset of a frequent $k$-itemset. Hence if any $(k-1)$-subset of a candidate $k$-itemset is not in $L_{k-1}$, then the candidate cannot be frequent either and so can be removed from $C_k$.

### 2.4 Illustration of Apriori Algorithm

Consider an example of Apriori, based on the following transaction database, D of Figure:1, with 9 transactions, to illustrate Apriori algorithm.

| TID | List of item_IDs |
|---|---|
| T100 | I1, I2, I5 |
| T200 | I2, I4 |
| T300 | I2, I3 |
| T400 | I1, I2, I4 |
| T500 | I1, I3 |
| T600 | I2, I3 |
| T700 | I1, I3 |
| T800 | I1, I2, I3, I5 |
| T900 | I1, I2, I3 |

**Figure: 1**

| Itemset | Sup. count |
|---|---|
| {I1} | 6 |
| {I2} | 7 |
| {I3} | 6 |
| {I4} | 2 |
| {I5} | 2 |

**Figure: 2** $C_1$

| Itemset | Sup. count |
|---|---|
| {I1} | 6 |
| {I2} | 7 |
| {I3} | 6 |
| {I4} | 2 |
| {I5} | 2 |

**Figure: 3** $L_1$

1. In the first iteration of the algorithm, each item is a member of the set of candidate 1-itemsets, $C_1$. The algorithm simply scans all of the transactions in order to count the number of occurrences of each item.
2. If minimum support count is set to 2, frequent 1-itemsets, $L_1$ , can then be determined from candidate 1-itemsets satisfying minimum support.
3. To discover the set of frequent 2-itemsets, $L_2$ , the algorithm uses $L_1 \bowtie L_1$ to generate a candidate set of 2-itemsets (Figure: 4).
4. The transactions D are scanned and the support count of each candidate itemset in $C_2$ is accumulated (Figure: 5).
5. The set of 2-itemsets, $L_2$ (Figure: 6), is then determined, consisting of those candidate 2-itemsets in $C_2$ having minimum support.

| Itemset | Itemset | Sup.count |
|---|---|---|
| {I1, I2} | {I1, I2} | 4 |
| {I1, I3} | {I1, I3} | 4 |
| {I1, I4} | {I1, I4} | 1 |
| {I1, I5} | {I1, I5} | 2 |
| {I2, I3} | {I2, I3} | 4 |
| {I2, I4} | {I2, I4} | 2 |
| {I2, I5} | {I2, I5} | 2 |
| {I3, I4} | {I3, I4} | 0 |
| {I3, I5} | {I3, I5} | 1 |
| {I4, I5} | {I4, I5} | 0 |

**Figure: 4** $C_2$     **Figure: 5** $C_2$

| Itemset | Sup.count |
|---|---|
| {I1, I2} | 4 |
| {I1, I3} | 4 |
| {I1, I5} | 2 |
| {I2, I3} | 4 |
| {I2, I4} | 2 |
| {I2, I5} | 2 |

**Figure: 6** $L_2$

6. The generation of the set of candidate 3-itemsets, $C_3$ , is detailed in Figure: 7 to Figure: 9. Here $C_3 = L_2 \bowtie L_2 = \{\{I1,I2,I3\}, \{I1,I2,I5\}, \{I1,I3,I5\}, \{I2,I3,I5\}, \{I2,I4,I5\}\}$. Based on the Apriori property that all subsets of a frequent itemset must also be frequent, the resultant candidate itemsets will be as in Figure: 7.

| Itemset | | Itemset | Sup. count |
|---|---|---|---|
| {I1, I2, I3} | | {I1, I2, I3} | 2 |
| {I1, I2, I5} | | {I1, I2, I5} | 2 |

**Figure: 7** $C_3$     **Figure: 8** $C_3$

| Itemset | Sup.count |
|---|---|
| {I1, I2, I3} | 2 |
| {I1, I2, I5} | 2 |

**Figure: 9** $L_3$

7. The transaction in D are scanned in order to determine $L_3$, consisting of those candidate 3-itemsets in $C_3$ having minimum support (Figure: 9).
8. The algorithm uses $L_3 | L_3$ to generate a candidate set of 4-itemsets, $C_4$. Although the join results in {{I1,I2,I3,I5}}, this itemset is pruned since its subset {{I2,I3,I5}} is not frequent. Thus, $C_4$ = {}, and the algorithm terminates.

## 2.5 Implementation of Association Rule on Text Data

Let us consider a set of transaction where each document is considered as a transaction as follows:
**1.** algorithm, network, graph, multicast, processor, system, parallel
**2.** cluster, network, design, message, processor, system, framework
**3.** algorithm, software, graph, method, session, analysis, parallel
**4.** switch, load, design, power, path, system, timing
**5.** cable, load, energy, power, current, motor, signal

After implementation of the Association rule (*considering minimum support as 0.4 & confidence 1*) we will get,
**a.** {**algorithm, graph**} $\Rightarrow$ {**parallel**} *from 1, 3*
**b.** {**network, processor**}$\Rightarrow${**system**} *from 1, 2*
**c.** {**design**} $\Rightarrow$ {**system**}   *from 2, 4*
**d.** {**load**} $\Rightarrow$ {**power**}   *from 4, 5*

## 3.1 Abstracts as Training Data

Abstracts from different thesis, research papers are considered as training document for developing a model for classifying new documents of unknown class. Most of the papers are collected from World Wide Web. Three categories of papers from Computer Science, Electrical and Electronic Engineering and Mechanical Engineering are considered as training documents.

## 3.2 Data Assumption, Consideration and Cleaning

Each abstract is considered as a Transaction in the Text data. So number of abstracts is equal to the number of transactions in the Transaction set (Text data). The next step is to clean the text data by removing unnecessary words. It is obvious that in a text document only few words can be termed as keywords, characterize the document. Unlike considering all words in a text, in our thesis work we have considered only those words that are related to the subject of the text. Some filtering process is adopted in order to remove unnecessary words in many text retrieval, text classification, and keyword extraction processes. We have followed a procedure which is similar to those conventional processes for filtering text data and collecting subject related words or keywords.

First, all stop words in addition to periods, commas, and punctuations from the text are removed. Second, we delete all words other than frequent words. We define a word as frequent if it occurs more than once in a text. For counting a word whether it is frequent or not, we assume singular and plural form of a word as same and keeping the singular form in the text. Finally, the remaining frequent words are considered as a single transaction data in the set of database transaction. This process is applied to all text data (abstracts) before applying association mining to the transaction database.

## 4. Deriving Associated word set from Training data

In this paper, total 115 numbers of abstracts (Mitchell, 1997, www) are used as training data for learning to classify text from all three categories, of which 47 are from Computer Science, 48 are from Electrical and Electronic Engineering and the rest 20 are from Mechanical Engineering papers. After preprocessing the text data association rule mining is applied to the set of transaction data where each frequent word set from each abstract is considered as a single transaction.

A partial list of generated large word set with their occurrence frequency in corresponding categories of Computer Science, Electrical and Electronic and Mechanical Engineering is given below in Table 4.1. The term large is used here because any subset (items more than one) of the frequent word set is also frequent according to the property of Apriori algorithm and therefore is not mentioned in the list. The support and confidence is set to 0.02 and 0.75 accordingly.

From the generated word set after applying association mining on training data we have found the following information based on the result.

Total No. of Word Set = 107
Total No. of Word Set from Computer Science = 43
Total No. of Word Set from Electrical & Electronic = 47
Total No. of Word Set from Mechanical = 17

Now we can recall the Naïve Bayes classifier for probability calculation.

$$\upsilon_{NB} = \text{argmax } P(\upsilon_j) \prod P(a_i | \upsilon_j)$$

The calculation for first term is based on the fraction of each target class in the training data.
Prior probability for Computer Science = 0.402
Prior probability for Electrical & Electrical = 0.44
Prior probability for Mechanical = 0.16

Then the second term of the equation is calculated by the following equation after adopting *m*-estimate approach [7] in order to avoid zero probability value,

$$\frac{n_k + 1}{n + |vocabulary|} \quad \ldots\ldots\ldots \text{(D)}$$

where,
$n$ = Total no of word set position in all training examples whose target value is $\upsilon_j$
$n_k$ = No. of times the word set found among all the training examples whose target value is $\upsilon_j$
$|vocabulary|$ = The total number of distinct word set found within all the training data

Replacing values for each category from Table 4.1 to equation (D) we will get probability values for each word set. The probability values for some of the word set is listed below in Table 4.2

## 5.1 Applying naïve Bayes theorem in Classification

Before classifying a new document the text data (abstract), target class of which is to be determined, is again preprocessed similar to the process applied to training data. The steps for preprocessing and classifying a new document can be summarized as follows:
* Remove periods, commas, punctuation, stop words. Collect words that have occurrence frequency more than once in the document.
* View the frequent words as word sets.
* Search for matching word set(s) or its subset (containing items more than one) in the list of word sets collected from training data with that of subset(s) (containing items more than one) of frequent word set of new document.
* Collect the corresponding probability values of matched word set(s) for each target class.
* Calculate the probability values for each target class from Naïve Bayes classification theorem.

Following the steps mentioned above, we can determine the target class of a new document. We will show an example document in the next section and classify it according to the steps described in section 5.1 and 5.2.

## 5.2 Example: Classifying a new document

Consider the following text (abstract) which can be any one of the categories of Computer Science, Electrical and Electronic Engg. or Mechanical Engg.

**Example**

*This paper discusses feedback control problems like regularization, noninteraction and linearization, for affine nonlinear singular systems. First, based on the constrained dynamic algorithm in affine nonlinear systems, an algorithm is introduced. By using such an algorithm, sufficient and necessary conditions are derived for the solvability of regularization problem. Then, another algorithm is proposed, based on which a sequence of integers can be defined for the system. It is shown that under some mild conditions, the dynamic part of singular systems can be linearized by using a regular feedback. Finally, an example is provided to illustrate the main results.*

After preprocessing the above text we have found the following Frequent words:
**{ feedback, problem, regularization, affine, nonlinear, singular, system, based, dynamic, algorithm, using, condition }**

Now search for word set(s) or its subset(s) from the list of word sets in Table: 5.2 matching with subset of frequent word set of new document. The following probability values in different categories are found accordingly.

| Matched Word Set from Training data | CS | EE | ME |
|---|---|---|---|
| {**problem,** graph, **algorithm** } | .027 | .0067 | .0067 |
| { irregular, multicast, **algorithm, system** } | 0.02 | 0.0067 | 0.0067 |
| {**algorithm,** message-passing, multicast, **system** } | 0.02 | 0.0067 | 0.0067 |
| {**dynamic, system,** interaction} | 0.0065 | 0.019 | 0.0065 |

| Large Word set found | Number of Occurrence in Documents | | |
|---|---|---|---|
| | CS | EE | ME |
| graph, algorithm | 5 | | |
| technology, processor, system | 4 | | |
| design, system | 4 | | |
| message-passing, system | 4 | | |
| oscillation, system, power, model | | 3 | |
| distribution, load, feeder, system | | 3 | |
| multicast, message-passing, system | 3 | | |
| destination, multicast, approach | 3 | | |
| System, result, model | | 3 | |
| power, control, system | | 3 | |
| problem, graph, algorithm | 3 | | |
| message, communication, system | 3 | | |
| stability, system, power | | 3 | |
| multidestination, message-passing, system | 3 | | |
| customer, feeder | | 3 | |
| instability, experiment | | | 3 |
| virtual, routing | 3 | | |
| device, power | | 3 | |
| block, power | | 3 | |
| voltage, power | | 3 | |
| shear, stress | | | 3 |
| generator, test | | 3 | |
| current, signal | | 3 | |
| stability, control, system, power, model, strategy, device, oscillation | | 2 | |
| Change, distribution, system, load, customer, temperature, feeder | | 2 | |
| pinout, framework, processor, technology, system, design | 2 | | |
| approach, message-passing, multicast, destination, system | 2 | | |
| broadcast, message, multicast, approach, destination | 2 | | |
| distribution, power, system, load, feeder | | 2 | |
| multidestination, communication, message, system, message-passing | 2 | | |
| power, damping, model, oscillation, system | | 2 | |
| irregular, multicast, algorithm, system | 2 | | |
| algorithm, message-passing, multicast, system | 2 | | |
| effect, system, power, load | | 2 | |
| multicast, network, message, algorithm | 2 | | |
| shear, experiment, rate, stress | | | 2 |
| sequential, generator, circuit, test | | 2 | |

Table 4.1: Word set with occurrence frequency

{multidestination, **based,** multicast, **system** }      0.02    0.0067    0.0067
**{using,** parameter, **system }**  0.0065  0.019   0.0065
**{ condition, algorithm }**         0.02    0.0067    0.0067

Matching subsets from frequent words of new document to be considered for probability calculation are:

1. **{algorithm,problem}**      2. **{algorithm,system}**
3. **{dynamic,system}**          4. **{system,based}**
5. **{system,using}**   6. **{algorithm,condition}**

| Large WordSet | CS | EE | ME |
|---|---|---|---|
| graph, algorithm | 0.04 | 0.0067 | 0.0067 |
| technology, processor, system | 0.033 | 0.0067 | 0.0067 |
| design, system | 0.033 | 0.0067 | 0.0067 |
| message-passing, system | 0.033 | 0.0067 | 0.0067 |
| oscillation, system, power, model | 0.0065 | 0.026 | 0.0065 |
| distribution, load, feeder, system | 0.0065 | 0.026 | 0.0065 |
| multicast, message-passing, system | 0.027 | 0.0067 | 0.0067 |
| destination, multicast, approach | 0.027 | 0.0067 | 0.0067 |
| system, result, model | 0.0065 | 0.026 | 0.0065 |
| power, control, system | 0.0065 | 0.026 | 0.0065 |
| problem, graph, algorithm | 0.027 | 0.0067 | 0.0067 |
| message, communication, system | 0.027 | 0.0067 | 0.0067 |
| stability, system, power | 0.0065 | 0.026 | 0.0065 |
| multidestination, message-passing, system | 0.027 | 0.0067 | 0.0067 |
| customer, feeder | 0.0065 | 0.026 | 0.0065 |
| instability, experiment | 0.0079 | 0.0079 | 0.031 |
| virtual, routing | 0.027 | 0.0067 | 0.0067 |
| device, power | 0.0065 | 0.026 | 0.0065 |
| block, power | 0.0065 | 0.026 | 0.0065 |
| voltage, power | 0.0065 | 0.026 | 0.0065 |
| shear, stress | 0.0079 | 0.0079 | 0.031 |
| generator, test | 0.0065 | 0.026 | 0.0065 |
| current, signal | 0.0065 | 0.026 | 0.0065 |
| stability, control, system, power, model, strategy, device, oscillation | 0.0065 | 0.019 | 0.0065 |
| Change, distribution, system, load, customer, temperature, feeder | 0.0065 | 0.019 | 0.0065 |
| pinout, framework, processor, technology, system, design | 0.02 | 0.0067 | 0.0067 |
| approach, message-passing, multicast, destination, system | 0.02 | 0.0067 | 0.0067 |
| broadcast, message, multicast, approach, destination | 0.02 | 0.0067 | 0.0067 |
| distribution, power, system, load, feeder | 0.0065 | 0.019 | 0.0065 |

Table 4.2: Word set with probability value{

The prior probability and probability values of word sets calculated using Naïve Bayes equation are:
Prior probability $P$(CS) = **0.40**, $P$(EE) = **0.44**,
$P$(ME) = **0.16**
$P$({algorithm,problem} | CS)=**0.027**,    $P$({algorithm,problem} | EE)=**0.0067**,
$P$({algorithm,problem} | ME)=**0.0067**
$P$({algorithm,condition} | CS)=**0.02**,   $P$({algorithm,condition} | EE)=**0.0067**,
$P$({algorithm,condition} | ME)=**0.0067**

*P*({algorithm,system} | CS)=**0.02**, *P*({algorithm,system} | EE)=**0.0067**, *P*({algorithm,system} | ME)=**0.0067**
*P*({dynamic,system} | CS)=**0.0065**, *P*({dynamic,system} | EE)=**0.019**, *P*({dynamic,system} | ME)=**0.0065**
*P*({based,system} | CS)=**0.02**, *P*({based,system} | EE)=**0.0067**, P({based,system} | ME)=**0.0067**
*P*({using,system} | CS)=**0.0065**, *P*({using,system} | EE)=**0.019**, *P*({using,system} | ME)=**0.0065**

*For* Computer Science = 0.4X0.027X0.027X0.02X0.0065X0.02X0.0065
= **0.00000000000492804**
*For* Electrical & Electronic = 0.44X0.0067X0.0067X0.0067X0.019X0.0067X0.019
= 0.0000000000000320080405964
*For* Mechanical = 0.16X0.0067X0.0067X0.0067X0.0065X0.0067X0.0065 = 0.0000000000000013622157796

From the above result we found the document classified as **Computer Science**.

### 5.3 Taking the effect of number of matching words

In the previous examples we consider only the probability values of word sets and the number of matching words in a word set has no effect in calculation. But we can take the effect of number of matching words by multiplying the fraction of matched words to the probability values during the calculation of probability for each target class.

**Example**

*Given a connected graph G = (V; E) with n vertices and m edges, the distance between two vertices in G is the weight of the shortest path between them. A subgraph G0 is a t-spanner (an approximate t-spanner) of G if, for every u, v 2 V, the distance between u and v in G0 is at most t (f(t)) times longer than the distance in G, where f(t) is a polynomial function of variable t and t <= f(t) < n. In this paper parallel algorithms for finding approximate t-spanners on both unweighted graphs and weighted graphs are given. If G is an unweighted graph, our algorithm requires O( ntk log n) time and M(n) processors, and the spanner generated has size of O(( ntk )1+1=t +n) and factor of O(tk+1); otherwise our algorithm requires O(( ntk )2 + (ntk)1 + 2 = (t?2) log n) time and O(n2) processors.*

After preprocessing the above text we have found the following Frequent words:
**{graph, vertices, distance, t-spanner, approximate, time, algorithm, unweighted, require, log, processor}**

Now search for word set(s) or its subset(s) from the list of word sets in Table: 5.2 matching with subset of frequent word set of new document. The following probability values in different categories are found accordingly.

**Matched Word Set**
**from Training data      CS      EE      ME**
{graph, algorithm}         0.04    0.0067   0.0067
{problem,**graph,algorithm**}0.027  0.0067   0.0067
{**time,**bound, **algorithm**}0.02   0.0067   0.0067

Matching subsets from frequent words of new document to be considered for probability calculation are:
1. **{algorithm, graph}** 2.**{algorithm, time}**

Therefore two-third (2/3) of the word sets *{problem, graph, algorithm}* & *{time, bound, algorithm}* matched with the subset of frequent words 1 & 2. The prior probability and probability values of word sets taking the effect of the fraction of matched word sets using naïve Bayes equation are:

Prior probability *P*(CS) = **0.40**; *P*(EE) = **0.44**; *P*(ME) = **0.16**
*P*({algorithm,graph} | CS) = **0.04** & **0.027*2/3**;
*P*({algorithm,graph} | EE) = **0.0067** & **0.0067*2/3**;
*P*({algorithm,graph} | ME) = **0.0067** & **0.0067*2/3** *P*({algorithm,time} | CS)=**0.02*2/3**;
*P*({algorithm,time} | EE) = **0.0067*2/3**;
*P*({algorithm,time} | ME) = **0.0067*2/3**
*For* Computer Science = 0.4X0.04X0.027 X2/3X0.2 X2/3 = **0.000003878496**
*For* Electrical & Electronic= 0.44X0.0067 X0.0067X2/3X0.0067X2/3= 0.000000059405504708
*For* Mechanical= 0.16X0.0067X0.0067 X2/3X0.0067X2/3 = 0.000000021602001712

### 6.1 Comparison with naïve Bayes classification

- Word Set of items two (at least) or more is generated from Association mining. So there is no option for considering a single word using association concept.
- Association mining largely reduces the number of words to be considered for    classifying texts,

keeping only words having association between them.
- Possibility of words common in more than one target classes is higher than the possibility of word set in more than one target classes. So considering a single word for classification increases the possibility of wrong classification.
- Considering word set instead of word for text classification increases the possibility of failure of text classification. But this possibility of failure can be reduced by considering increased number of training data. For example, we can consider the following frequent words collected after preprocessing an abstract for classification.

### 6.2 Efficiency of classifying a text

In this work, classifying a new document depends on the associated word sets generated from training documents. So the number of training documents is vital in generating the number of word sets used to determine the class of a new document. The greater number of word sets from training documents reduces the possibility of failure to classify a new document. We have considered only a total 115 number of documents as training data (Mitchell, 1997) which is very few and insufficient compared to Naive Bayes example of text classification where 20,000 documents taken for developing the learning system and that system gives 89% efficiency in text classification.

We have started with 60 documents (20+20+20) initially then increase the number to 115. Increase in doubled the number of documents also doubled the number of generated word sets, which in turn increases significantly its ability to classify a text.

### 6.3 Limitations and future work

In our training set of data, although all the abstracts have almost equal size in length, they have slightly different number of frequent words after preprocessing them. In order to avoid null attribute value in any transaction in the set of transaction database, we have considered 13(thirteen) frequent words from each text. The reason is that, null attribute values in the transaction set produce word sets containing null values. These word sets containing null values have no use in classification.

Texts with less than 13 frequent words are discarded (remaining 115 documents) and are not considered as training data. A process is followed for selecting 13 frequent words from documents having frequent words more than 13 based on occurrence frequency and position of frequent words from the beginning of a text in case of same frequency words. In other words, higher frequency words are considered first, then for the same frequency words, word that occurs earlier from the beginning of the text gets priority for selection over others.

Considering increased number of attributes for generating associated word sets, increase the possibility of generating greater number of words in a word set and also increase the total number of word sets.